\documentclass[conference]{IEEEtran}
\IEEEoverridecommandlockouts
\usepackage{cite}
\usepackage{amsmath,amssymb,amsfonts}
\usepackage{algorithmic}
\usepackage{graphicx}
\usepackage{textcomp}
\usepackage{xcolor}
\usepackage{float}
\usepackage{placeins}
\def\BibTeX{{\rm B\kern-.05em{\sc i\kern-.025em b}\kern-.08em
    T\kern-.1667em\lower.7ex\hbox{E}\kern-.125emX}}
\begin{document}

\title{Conversion and Implementation of State-of-the-Art Deep Learning Algorithms for the Classification of Diabetic Retinopathy\\
}

\author{\IEEEauthorblockN{1\textsuperscript{st} Mihir Rao\\
\textit{Chatham High School}\\
Chatham, New Jersey, USA \\
mihirraov@gmail.com}
\and
\IEEEauthorblockN{2\textsuperscript{nd} Michelle Zhu}
\IEEEauthorblockA{\textit{Department of Computer Science} \\
\textit{Montclair State University}\\
Montclair, New Jersey, USA \\
zhumi@montclair.edu}
\and
\IEEEauthorblockN{3\textsuperscript{rd} Tianyang Wang}
\IEEEauthorblockA{\textit{Department of Computer Science} \\
\textit{Austin Peay State University}\\
Tennessee, USA \\
toseattle@siu.edu}
}

\maketitle

\begin{abstract}
Diabetic retinopathy (DR) is a retinal microvascular condition that emerges in diabetic patients. DR will continue to be a leading cause of blindness worldwide, with a predicted 191.0 million globally diagnosed patients in 2030. Microaneurysms, hemorrhages, exudates, and cotton wool spots are common signs of DR. However, they can be small and hard for human eyes to detect. Early detection of DR is crucial for effective clinical treatment. Existing methods to classify images require much time for feature extraction and selection, and are limited in their performance. Convolutional Neural Networks (CNNs), as an emerging deep learning (DL) method, have proven their potential in image classification tasks. In this paper, comprehensive experimental studies of implementing state-of-the-art CNNs for the detection and classification of DR are conducted in order to determine the top performing classifiers for the task. Five CNN classifiers, namely Inception-V3, VGG19, VGG16, ResNet50, and InceptionResNetV2, are evaluated through experiments. They categorize medical images into five different classes based on DR severity. Data augmentation and transfer learning techniques are applied since annotated medical images are limited and imbalanced. Experimental results indicate that the ResNet50 classifier has top performance for binary classification and that the InceptionResNetV2 classifier has top performance for multi-class DR classification.
\end{abstract}


\section{Introduction}

Diabetic retinopathy (DR) is a retinal microvascular condition that emerges as a direct result of diabetes. High blood sugar levels allow glucose to block blood vessels, in this case in the retina \cite{b1}. This leads to microaneurysms, which are swollen sections of blood vessels in the retina. When these microaneurysms leak, they are called hemorrhages \cite{b2}. These hemorrhages allow cotton wool spots to form, which are accumulations of axoplasmic material in the back of the eye, along with exudates \cite{b3}. In order to effectively treat DR, it must be detected in its early stages. However, most people with the condition are unaware of the fact that they must have their vision examined often, thus allowing the condition to pass the early stages and into the later stages undetected \cite{b4}. Additionally, DR patients in resource-poor countries lack effective DR identification technology and clinicians in order to make official diagnoses and treatment plans \cite{b4}. This means that not only must DR be detected in its early stages, but the detection technology must be easily accessible for people who do not have access to eye specialists and adequate technology.

In 2030, it is estimated that there will be 191.0 million people with DR globally. This is approximately a 50\% jump from the 126.6 million people with DR globally in 2010. Of the 191.0 million people, 56.3 million people are expected to have vision-threatening diabetic retinopathy (VTDR) if action is not taken \cite{b5}. In the United States alone, the number of Americans aged 40 and older in 2050 with DR is predicted to be 16.0 million people, while the number of people with VTDR is expected to be 3.4 million. These numbers are approximately three times the amount of people when compared to 2005 when there were 5.5 million people with DR and 1.2 million people with VTDR \cite{b6}. Clearly, early and accurate DR detection is not only vital in the present day, but it will continue to be necessary for decades to come.

Many medical imaging techniques such as computed tomography (CT) and magnetic resonance imaging (MRI) have become indispensable tools in clinical research and diagnosis. Classifying medical images has been playing an important role in disease diagnosis and medical treatment \cite{b7}. 

At present, the most widely used method for the detection of diabetic retinopathy is a retinal eye exam \cite{b8}. This approach involves an eye specialist looking through a patient’s pupil and at the back of their eye. The specialist looks for some of the most common symptoms of the disease, including microaneurysms, hemorrhages, exudates, and cotton wool spots. Additionally, some detection methods involve using fundus photography to take a picture of a patient’s retina, allowing an eye specialist to conduct the same examination by looking at a retinal image on a computer screen. Lack of ophthalmologists will leave a large portion of patients undiagnosed. In addition, human errors are unavoidable. 

Due to the vast amount of medical images and human fatigue as well as errors, relying on professional ophthalmologists will be very expensive and inefficient. Thus, machine learning approaches have been used for this purpose. Medical image classification can generally be categorized into supervised and unsupervised classification methods. Supervised methods require samples to be pre-annotated and include K-nearest neighbor algorithm, Bayesian models, logistic regression, neural networks, and support vector machines. Unsupervised methods automatically detect the similarity among samples and include K-means clustering, auto-encoders, and principal component analysis (PCA). 

In recent years, deep learning (DL) has increasingly attracted researchers’ attention for medical image classification. DL is a machine learning technique in which neural networks are trained on collections of data in order to learn patterns and extract features from them. Trained DL models can then be used to predict certain details about unseen test data, making them useful tools for medical image classification. Specifically, CNNs are DL models that have been proven effective for feature extraction and pattern recognition in image data, especially in a medical context. 

Deep learning integrates supervised methods with unsupervised methods, and some image classification experiments using convolutional neural networks (CNNs) achieve performance close to what a specialized physician can achieve \cite{b9}. 

In this paper, the aim is to detect diabetic retinopathy by classifying a retinal image into one of the five different levels (classes), as shown in Fig. 1, based on the disease severity. The medical images used in this paper are very limited and unbalanced due to the data privacy concern and labeling efforts. 

To address the data challenges in the medical imaging data, several techniques are used. Data augmentation technique is used to counteract the small data size by rotating and scaling the existing images. Noise is intentionally introduced to increase the noise tolerance level of the models. Transfer learning utilizes some pre-trained model as a base to build the new model. This allows pre-acquired weights to be propagated into the classification task. During experiments, the selected models are trained on a publicly available dataset \cite{b10}.

\begin{figure*}[htbp]
\centerline{\includegraphics[scale=0.8]{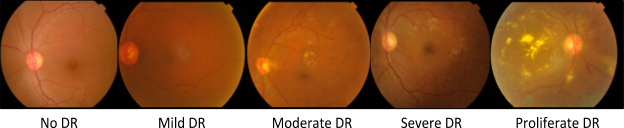}}
\caption{Retinal images showing the progression of DR.}
\label{fig}
\end{figure*}

\section{Related Work}

It has been proven that CNNs can automatically extract more distinct and effective features than handcrafted feature extraction methods. Deep learning methods usually outperform traditional machine learning methods, such as SVMs (support vector machines), since the SVM method is designed for a small sample size and is not suitable for large samples \cite{b9} \cite{b11}. Deep learning networks are widely adopted in order to improve the image classification performance since 2012 \cite{b12}. In particular, Krizhevsky et al. used a CNN to classify 1.2 million images into 1000 classes, with benchmark performance at the ImageNet Large Scale Visual Recognition Challenge (ILSVRC) 2012 \cite{b13}. 

In order to improve the accuracy, researchers have been working to add more layers to the CNNs for large-scale image classification. VGG neural networks address the depth of CNNs and support up to 19 layers \cite{b14}. Very small 3×3 filters in all convolutional layers are used to reduce the number of parameters. Additionally, it has been demonstrated that 3×3 filters are most effective according to covariance analysis \cite{b15}. Significant performance gain has been observed by increasing the depth when compared to previous architectures \cite{b14}.

However, simply adding layers to CNNs for higher accuracy causes complications such as overfitting, degradation and computing and memory burden. Skip connections among the layers are proposed in residual networks, which learn an additive residual function with respect to an identity mapping derived from the preceding layers inputs \cite{b16}. The residual architectures are capable of better fusing features and thus address the issue of gradient explosion or vanishing.

The large number of parameters and large data size contribute to the success of CNN models. However, training deep networks usually has high demands for high performance computing resources, such as powerful GPUs and efficient storage systems. Multiple-GPUs have been used to speed up training. The data parallelism can be exploited by dividing each batch of training images into several smaller batches, computed in parallel on each GPU. For example, Simonyan and Zisserman trained their VGG nets of 144M parameters on four NVIDIA Titan Black GPUs \cite{b14}. 

Model-based transfer learning for the neural network contains two stages, namely network pre-training with benchmark datasets, such as ImageNet \cite{b17}, and fine-tuning the pre-trained networks with specific target datasets. The two-stage method has been very popular in tasks involving medical images due to the limited sizes of specific target datasets. The pre-trained networks can capture some general features from similar benchmark images and these features can be further fine-tuned in the second stage. Experiments show that feature reuse primarily happens in the lowest layers \cite{b18}.

\section{Methods}

\subsection{Addressing Class Imbalance in the Dataset}

A Kaggle dataset titled APTOS 2019 Blindness Detection (APTOS stands for Asia Pacific Tele-Ophthalmology Society) was used to train and test models \cite{b10}. The dataset consists of 3662 retinal images across five different stages of diabetic retinopathy (DR): no DR, mild DR, moderate DR, severe DR, and proliferate DR. These classes are annotated as values 0 through 4. As shown in Fig. 2, classes of images were grouped together and re-annotated based on the classification task at hand (binary, 3-class, or 5-class). Class imbalance was addressed by oversampling images in order to achieve a more uniform distribution of images across classes. Additionally, oversampled images were randomly rotated, reflected, and noisified in order to prevent training models on duplicate data and to increase the overall noise resistance of the system.

\begin{figure}[!h]
\centerline{\includegraphics[scale=0.8]{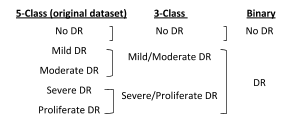}}
\caption{Diagram showing the approach taken for image grouping and reannotation based on the classification task at hand.}
\label{fig}
\end{figure}

\subsection{Image Pre-Processing}

The original dataset used in this project consists of images of vastly varying dimensions and characteristics, including empty space around the actual retina in the image and the brightness of the image as a whole. In order to feed such data into a CNN, the images need to be pre-processed. This was done in a series of steps. Firstly, by checking for pixels throughout the images that are completely black, the regions of empty space were able to be detected and then cropped out. Then, in an effort to help normalize the brightness of the images as well as to help bring out some of the important features of the images, weighted arrays consisting of a Gaussian blurred versions of the images were added to their corresponding resized images. Thirdly, the images were circle cropped, with the center of the circle lying at the center of the image and the circumference of the circle touching the edges of the image. Not only did this allow for excess space around the retina to be further eliminated, but it also led to more uniformity throughout the dataset. Lastly, the images were resized to a common dimension of 512x512 pixels. This specific dimension was chosen as a starting point since the raw images in the dataset had an average size of approximately 1527x2015 pixels. The smaller the image is resized, the more the details from the raw image are lost. So, resizing the images to 512 square pixels allowed for high image resolution to be maintained while achieving time and memory efficiency during training. Fig. 3 shows two examples of image pre-processing conducted on raw retinal images.

\begin{figure}[!b]
\centerline{\includegraphics[scale=0.85]{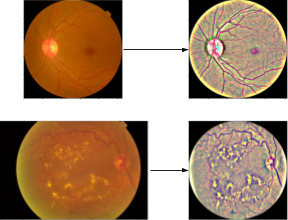}}
\caption{Two examples of raw retinal images undergoing image pre-processing.}
\label{fig}
\end{figure}

\subsection{Transfer Learning}

Unlike training from scratch, transfer learning aims to transfer the knowledge, which was learned from another dataset, to a target problem. These models are repurposed, as their weights and biases are from their initial training. These parameters may still help with some high-level feature extraction due to certain level of relevance between the two models. Additionally, the selected pre-trained models have been proven successful in other classification problems, further making them good candidates for the problem of DR detection and classification \cite{b14}\cite{b19}\cite{b20}. During experiments, the following pre-trained models were selected for further fine-tuning: ResNet50, VGG16, VGG19, Inception-V3, and InceptionResNetV2. Since these models were initially trained for tasks that categorized images into a large number of unique classes, the final layer of each model did not match the required architecture for binary, 3-class, and 5-class classification tasks. In order to address this, a series of additional layers were added to the end of each classifier in order to fit the classification task at hand: a flatten layer to reduce model output to a 1-dimensional space, a series of fully-connected dense layers, and a final dense layer with 1 node for the binary task, 3 nodes for the 3-class task, and 5 nodes for the 5-class task, respectively.

\section{Experiments and Results}

\subsection{Experimental Setups}

For the binary classification task, various models were tested across different optimization techniques. Specifically, the models that were tested were ResNet50, VGG16, VGG19, and Inception-V3. All of these models were tested across two optimizers: Adam and Stochastic Gradient Descent (SGD). A learning rate of 0.001 was used for both optimizers across all experiments. The final layer of all of the classifiers in the binary classification task had a sigmoid activation function implemented, providing outputs between 0 and 1. In differentiating between outputs of the negative and positive class, a threshold value of <=0.5 was used for the negative class while a value of >0.5 was used for the positive class. During training, the base transfer learning model’s layers were frozen.

For the 3-class classification task, two phases of experiments were conducted. For the first phase, similar approaches were taken as for the binary classification task with respect to model architecture and optimizers. However, in the final layer of all 3-class classification models, a Softmax activation across 3 nodes was used, providing probabilities across the three classes. Again, a learning rate of 0.001 was used for both optimizers. In order to interpret the Softmax probabilities produced by the model as a prediction, the Argmax function was implemented, which returned the index of the element in the probability array with the highest value, thus returning the model’s most confident prediction for a certain image. Based on the results from this initial phase of experiments, adjustments were made and a second phase of experiments were conducted. The adjustments made were the learning rate being decreased by a factor of 10, the kernels of the entire model being initialized using the He Uniform initializer \cite{b16}, and the unfreezing of layers in the base transfer learning model. However, the adjustments yielded memory limitations for model training of ResNet50 and Inception-V3 due to their output tensors being extremely large once flattened in the classifier (524288 values for ResNet50 and 401408 values for Inception-V3). This is several fold higher than the number of output parameters by the VGG variants (131072 values). So, the VGG variants were trained using the above-mentioned method while ResNet50 and Inception-V3 training occurred on images of size 224x224 pixels and 299x299 pixels, respectively. This helped reduce the output tensor from 4 to 2 dimensions, thus alleviating the memory limitation. 224 and 299 square pixels is the default input size for the ResNet50 and Inception-V3 trained on ImageNet, so feeding this size into ResNet50 and Inception-V3 allowed for training on more default settings. In order to collect more thorough results, training was conducted for the VGG variants on 224x224 pixel images for comparison purposes. It is recommended that the ResNet50 and Inception-V3 models be trained just like the VGG variants (on 512x512 pixel images) as part of future work when adequate resources are available. Once phase 2 of results were analyzed, based on the relatively good performance of ResNet50 and Inception-V3, the InceptionResNetV2 architecture, also known as Inception-V4, was trained with the Adam optimizer using the same training parameters as used with the other models. Inception-V4 takes the Inception-V3 architecture and incorporates residual connection much like those in the ResNet variants \cite{b19}. The input image size for this model was 299x299 pixels. The results of testing this model are presented with the phase 2 experimentation results.

For the 5-class classification task, just like with the 3-class classification task, two phases of experiments were conducted. The first phase conducted the same experiments that were conducted in the phase 1 of experimentation for the 3-class classification, only the softmax activation on the final layer of the classifier was modified to fit the 5-class classification task. Based on the results of this initial phase, a second phase of experimentation was conducted after making adjustments to training parameters. These adjustments were the same as those made prior to the second phase of the 3-class classification task. Additionally, the second phase for the 5-class classification task used the same experimental setups as those used in phase 2 for the 3-class classification task. Based on the observed performance of the ResNet50 and Inception-V3 models, just like for the 3-class classification task, the InceptionResNetV2 model was experimented with both the Adam and SGD optimizers. Again, the same training parameters were used for this model as used for the 3-class classification task. The results of testing this model are presented with the phase 2 experimentation results for the 5-class classification task.

For all classification types, test sets were created using random 20\% samples and validation sets were random 20\% samples of the train set. During training, an early-stopping callback was implemented. This would monitor validation accuracy during training and stop training once the validation accuracy began to decrease, indicating overfitting. The callback would then restore the model’s best weights from the penultimate epoch. Additionally, across all binary task experiments and experiments in phase 1 of the multi-class approaches, weight initialization for the base model was done using the ImageNet weights provided for the model being tested. These provided weights enable the model with high-level feature extraction abilities.

\subsection{Model Performance Analysis}

Table I shows the collected results for the binary classification task. A confusion matrix in Fig. 4 and receiver-operator characteristic (ROC) curves in Fig. 5 are shown for the binary classification experimental setup that yielded the best results. It was found that the ResNet50 architecture accompanied with the Adam optimizer yielded the best results for the binary classification task, including an accuracy of 96.59\% when tested on unseen test data and micro and macro-average area-under-the-curve (AUC) values of 0.99.

\begin{table}[htbp]
\caption{Binary Classification Experimental Results}
\begin{center}
\begin{tabular}{|c|c|c|c|c|c|c|}
\hline
\textbf{}&\multicolumn{4}{|c|}{\textbf{Model and Optimizer}} \\
\cline{2-5} 
\textbf{}&\multicolumn{4}{|c|}{\textbf{Adam}} \\
\cline{2-5} 
\textbf{Metric} & \textbf{\textit{ResNet50}}& \textbf{\textit{VGG16}}& \textbf{\textit{VGG19}} & \textbf{\textit{Inception-V3}}\\
\hline
Test Accuracy&\textbf{0.9659}&0.9503&0.9517&0.8963\\
\hline
Precision&\textbf{0.97}&0.95&0.95&0.89\\
\hline
Recall&\textbf{0.97}&0.95&0.95&0.89\\
\hline
Micro Average AUC&\textbf{0.99}&0.97&0.98&0.94\\
\hline
Macro Average AUC&\textbf{0.99}&0.98&0.98&0.94\\
\hline
F1-Score&\textbf{0.9659}&0.9502&0.9487&0.892\\
\hline
\textbf{}&\multicolumn{4}{|c|}{\textbf{Stochastic Gradient Descent}} \\
\cline{2-5}
\textbf{Metric} & \textbf{\textit{ResNet50}}& \textbf{\textit{VGG16}}& \textbf{\textit{VGG19}} & \textbf{\textit{Inception-V3}}\\
\hline
Test Accuracy&0.956&0.9119&0.9219&0.7827\\
\hline
Precision&0.95&0.91&0.92&0.82\\
\hline
Recall&0.95&0.91&0.92&0.79\\
\hline
Micro Average AUC&\textbf{0.99}&0.97&0.97&0.88\\
\hline
Macro Average AUC&\textbf{0.99}&0.97&0.97&0.93\\
\hline
F1-Score&0.953&0.9117&0.9219&0.781\\
\hline
\end{tabular}
\label{tab1}
\end{center}
\end{table}

\begin{figure}[htbp]
\centerline{\includegraphics[scale=0.7]{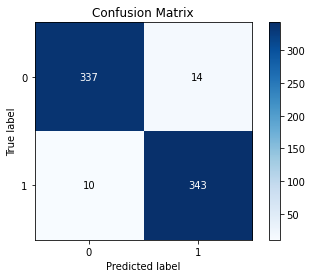}}
\caption{Confusion matrix for the testing results of the best binary classification model.}
\label{fig}
\end{figure}

\begin{figure}[htbp]
\centerline{\includegraphics[scale=0.6]{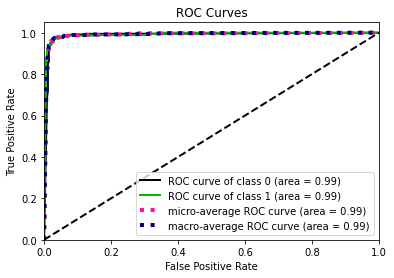}}
\caption{Receiver-operator characteristic (ROC) curves for the testing results of the best binary classifier.}
\label{fig}
\end{figure}

Table II shows the model testing accuracies for phase 1 of 3-class classification experimentation. It can be seen that the majority of the models are not able to pass roughly 84\% testing accuracy. This could be due to the fact that with 3-class classification, merely training the end classifier and not the base transfer learning model itself may not be sufficient to help the model differentiate between the mild/moderate and severe/proliferate classes. So, unfreezing of the base model layers, initializing weights in a specific manner, and decreasing the learning rate to 0.0001 was necessary. Table III shows the results for phase 2 of experimentation for 3-class classification with 512x512 pixel images. Table IV shows the results for phase 2 of experimentation for 3-class classification with 299x299 pixel images for the Inception-V3 and InceptionResNetV2 models and 224x224 pixel images for the other models. A confusion matrix in Fig. 6 and receiver-operator characteristic (ROC) curves in Fig. 7 are shown for the 3-class classification experimental setup that yielded the best results. It was found that the InceptionResNetV2 architecture accompanied with the Adam optimizer yielded the best results for the 3-class classification task, including an accuracy of 88.14\% and a micro and macro-average AUC values of 0.98 and 0.97, respectively.

\begin{table}[htbp]
\caption{3-Class Classification Phase 1 Test Accuracies}
\begin{center}
\begin{tabular}{|c|c|c|c|c|c|c|}
\hline
\textbf{}&\multicolumn{4}{|c|}{\textbf{Model and Optimizer}} \\
\cline{2-5} 
\textbf{}&\multicolumn{4}{|c|}{\textbf{Adam}} \\
\cline{2-5} 
\textbf{Metric} & \textbf{\textit{ResNet50}}& \textbf{\textit{VGG16}}& \textbf{\textit{VGG19}} & \textbf{\textit{Inception-V3}}\\
\hline
Test Accuracy&0.795&0.8088&0.8309&0.602\\
\hline
\textbf{}&\multicolumn{4}{|c|}{\textbf{Stochastic Gradient Descent}} \\
\cline{2-5}
\textbf{Metric} & \textbf{\textit{ResNet50}}& \textbf{\textit{VGG16}}& \textbf{\textit{VGG19}} & \textbf{\textit{Inception-V3}}\\
\hline
Test Accuracy&\textbf{0.8419}&0.6976&0.7491&0.5809\\
\hline
\end{tabular}
\label{tab1}
\end{center}
\end{table}

\begin{table}[htbp]
\caption{3-Class Classification Phase 2 Experimental Results on 512x512 Pixel Input Images}
\begin{center}
\begin{tabular}{|c|c|c|}
\hline
\textbf{}&\multicolumn{2}{|c|}{\textbf{Model and Optimizer}} \\
\cline{2-3} 
\textbf{}&\multicolumn{2}{|c|}{\textbf{Adam}} \\
\cline{2-3} 
\textbf{Metric}& \textbf{\textit{VGG16}}& \textbf{\textit{VGG19}}\\
\hline
Test Accuracy&0.7454&0.7528\\
\hline
Precision&0.74&0.77\\
\hline
Recall&0.75&0.76\\
\hline
Micro Average AUC&0.9&0.9\\
\hline
Macro Average AUC&0.88&0.89\\
\hline
F1-Score&0.7434&0.7616\\
\hline
\textbf{}&\multicolumn{2}{|c|}{\textbf{Stochastic Gradient Descent}} \\
\cline{2-3}
\textbf{Metric}& \textbf{\textit{VGG16}}& \textbf{\textit{VGG19}}\\
\hline
Test Accuracy&0.7472&\textbf{0.7932}\\
\hline
Precision&0.76&\textbf{0.81}\\
\hline
Recall&0.76&\textbf{0.8}\\
\hline
Micro Average AUC&0.9&\textbf{0.93}\\
\hline
Macro Average AUC&0.89&\textbf{0.92}\\
\hline
F1-Score&0.7563&\textbf{0.8063}\\
\hline
\end{tabular}
\label{tab1}
\end{center}
\end{table}

\begin{table*}[htbp]
\caption{3-Class Classification Phase 2 Experimental Results on 224x224 and 299x299 Pixel Input Images}
\begin{center}
\begin{tabular}{|c|c|c|c|c|c|c|}
\hline
\textbf{}&\multicolumn{5}{|c|}{\textbf{Model and Optimizer}} \\
\cline{2-6} 
\textbf{}&\multicolumn{5}{|c|}{\textbf{Adam}} \\
\cline{2-6} 
\textbf{Metric} & \textbf{\textit{ResNet50}}& \textbf{\textit{VGG16}}& \textbf{\textit{VGG19}} & \textbf{\textit{Inception-V3}}& \textbf{\textit{InceptionResNetV2}}\\
\hline
Test Accuracy&0.7849&0.3695&0.3548&0.8116&\textbf{0.8814}\\
\hline
Precision&0.79&0.12&0.12&0.83&\textbf{0.88}\\
\hline
Recall&0.78&0.33&0.33&0.81&\textbf{0.88}\\
\hline
Micro Average AUC&0.92&0.52&0.52&0.94&\textbf{0.98}\\
\hline
Macro Average AUC&0.9&0.5&0.5&0.94&\textbf{0.97}\\
\hline
F1-Score&0.7778&0.1772&0.1759&0.8046&\textbf{0.8827}\\
\hline
\textbf{}&\multicolumn{5}{|c|}{\textbf{Stochastic Gradient Descent}} \\
\cline{2-6}
\textbf{Metric} & \textbf{\textit{ResNet50}}& \textbf{\textit{VGG16}}& \textbf{\textit{VGG19}}& \textbf{\textit{Inception-V3}}& \textbf{\textit{InceptionResNetV2}}\\
\hline
Test Accuracy&0.3493&0.3419&0.3906&0.3125&0.4164\\
\hline
Precision&0.33&0.14&0.37&0.29&0.35\\
\hline
Recall&0.35&0.33&0.4&0.32&0.44\\
\hline
Micro Average AUC&0.49&0.47&0.55&0.5&0.52\\
\hline
Macro Average AUC&0.49&0.41&0.57&0.48&0.56\\
\hline
F1-Score&0.2413&0.1807&0.3483&0.2239&0.346\\
\hline
\end{tabular}
\label{tab1}
\end{center}
\end{table*}

\begin{figure}[htbp]
\centerline{\includegraphics[scale=0.7]{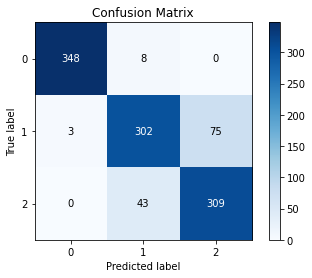}}
\caption{Confusion matrix for the testing results of the best 3-class classifier.}
\label{fig}
\end{figure}

\begin{figure}[htbp]
\centerline{\includegraphics[scale=0.6]{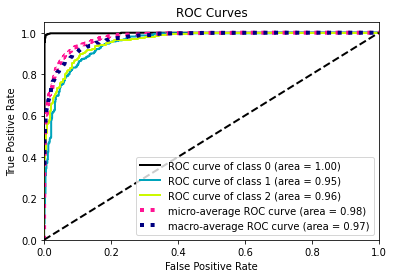}}
\caption{Receiver-operator characteristic (ROC) curves for the testing results of the best 3-class classifier.}
\label{fig}
\end{figure}

Table V shows the accuracies for the first phase of 5-class classification experimentation. It can be seen that the majority of the models are not able to pass roughly 70\% testing accuracy. This could be due to the fact that with non-multi-class approaches, merely training the end classifier and not the base transfer learning model itself yielded promising results. However, as the 5-class classification task requires more detailed classification by the model, especially between the mild and moderate classes and the severe and proliferate classes, the mentioned adjustments were made before phase 2 of experiments. Table VI shows the results for phase 2 of experimentation for 5-class classification with 512x512 pixel images. Table VII shows the results for phase 2 of experimentation for 5-class classification with 299x299 pixel images for the Inception-V3 and InceptionResNetV2 models and 224x224 pixel images for the other models. A confusion matrix in Fig. 8 and receiver-operator characteristic (ROC) curves in Fig. 9 are shown for the 5-class classification experimental setup that yielded the best results. It was found that the InceptionResNetV2 architecture accompanied with the Adam optimizer yielded the best results for the 5-class classification task, including an accuracy of 85.02\% and a micro and macro-average AUC values of 0.97.

\begin{table}[htbp]
\caption{5-Class Classification Phase 1 Test Accuracies}
\begin{center}
\begin{tabular}{|c|c|c|c|c|c|c|}
\hline
\textbf{}&\multicolumn{4}{|c|}{\textbf{Model and Optimizer}} \\
\cline{2-5} 
\textbf{}&\multicolumn{4}{|c|}{\textbf{Adam}} \\
\cline{2-5} 
\textbf{Metric} & \textbf{\textit{ResNet50}}& \textbf{\textit{VGG16}}& \textbf{\textit{VGG19}} & \textbf{\textit{Inception-V3}}\\
\hline
Test Accuracy&0.6374&\textbf{0.7037}&0.6643&0.2128\\
\hline
\textbf{}&\multicolumn{4}{|c|}{\textbf{Stochastic Gradient Descent}} \\
\cline{2-5}
\textbf{Metric} & \textbf{\textit{ResNet50}}& \textbf{\textit{VGG16}}& \textbf{\textit{VGG19}} & \textbf{\textit{Inception-V3}}\\
\hline
Test Accuracy&0.6681&0.6519&0.6288&0.1832\\
\hline
\end{tabular}
\label{tab1}
\end{center}
\end{table}

\begin{table}[htbp]
\caption{5-Class Classification Phase 2 Experimental Results on 512x512 Pixel Input Images}
\begin{center}
\begin{tabular}{|c|c|c|}
\hline
\textbf{}&\multicolumn{2}{|c|}{\textbf{Model and Optimizer}} \\
\cline{2-3} 
\textbf{}&\multicolumn{2}{|c|}{\textbf{Adam}} \\
\cline{2-3} 
\textbf{Metric}& \textbf{\textit{VGG16}}& \textbf{\textit{VGG19}}\\
\hline
Test Accuracy&\textbf{0.8066}&0.4547\\
\hline
Precision&\textbf{0.81}&0.41\\
\hline
Recall&\textbf{0.81}&0.45\\
\hline
Micro Average AUC&\textbf{0.96}&0.8\\
\hline
Macro Average AUC&\textbf{0.95}&0.77\\
\hline
F1-Score&\textbf{0.8106}&0.3752\\
\hline
\textbf{}&\multicolumn{2}{|c|}{\textbf{Stochastic Gradient Descent}} \\
\cline{2-3}
\textbf{Metric}& \textbf{\textit{VGG16}}& \textbf{\textit{VGG19}}\\
\hline
Test Accuracy&0.6853&0.7295\\
\hline
Precision&0.7&0.72\\
\hline
Recall&0.7&0.72\\
\hline
Micro Average AUC&0.92&0.93\\
\hline
Macro Average AUC&0.91&0.92\\
\hline
F1-Score&0.6931&0.7199\\
\hline
\end{tabular}
\label{tab1}
\end{center}
\end{table}

\begin{table*}[htbp]
\caption{5-Class Classification Phase 2 Experimental Results on 224x224 and 299x299 Pixel Input Images}
\begin{center}
\begin{tabular}{|c|c|c|c|c|c|c|}
\hline
\textbf{}&\multicolumn{5}{|c|}{\textbf{Model and Optimizer}} \\
\cline{2-6} 
\textbf{}&\multicolumn{5}{|c|}{\textbf{Adam}} \\
\cline{2-6} 
\textbf{Metric} & \textbf{\textit{ResNet50}}& \textbf{\textit{VGG16}}& \textbf{\textit{VGG19}} & \textbf{\textit{Inception-V3}}& \textbf{\textit{InceptionResNetV2}}\\
\hline
Test Accuracy&0.7893&0.2091&0.2042&0.6228&\textbf{0.8502}\\
\hline
Precision&0.8&0.04&0.04&0.55&\textbf{0.85}\\
\hline
Recall&0.79&0.2&0.2&0.63&\textbf{0.85}\\
\hline
Micro Average AUC&0.95&0.5&0.5&0.91&\textbf{0.97}\\
\hline
Macro Average AUC&0.94&0.5&0.5&0.88&\textbf{0.97}\\
\hline
F1-Score&0.7905&0.0692&0.0677&0.5746&\textbf{0.8495}\\
\hline
\textbf{}&\multicolumn{5}{|c|}{\textbf{Stochastic Gradient Descent}} \\
\cline{2-6}
\textbf{Metric} & \textbf{\textit{ResNet50}}& \textbf{\textit{VGG16}}& \textbf{\textit{VGG19}}& \textbf{\textit{Inception-V3}}& \textbf{\textit{InceptionResNetV2}}\\
\hline
Test Accuracy&0.222&0.2328&0.2548&0.1827&0.1994\\
\hline
Precision&0.35&0.16&0.38&0.18&0.13\\
\hline
Recall&0.21&0.24&0.25&0.19&0.19\\
\hline
Micro Average AUC&0.49&0.49&0.57&0.48&0.52\\
\hline
Macro Average AUC&0.5&0.49&0.53&0.47&0.55\\
\hline
F1-Score&0.1517&0.1595&0.1788&0.1042&0.1435\\
\hline
\end{tabular}
\label{tab1}
\end{center}
\end{table*}

\begin{figure}[htbp]
\centerline{\includegraphics[scale=0.7]{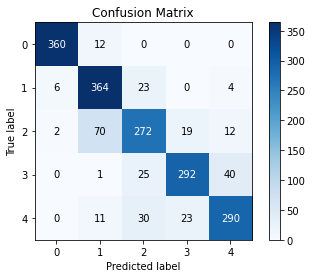}}
\caption{Confusion matrix for the testing results of the best 5-class classifier.}
\label{fig}
\end{figure}

\begin{figure}[htbp]
\centerline{\includegraphics[scale=0.6]{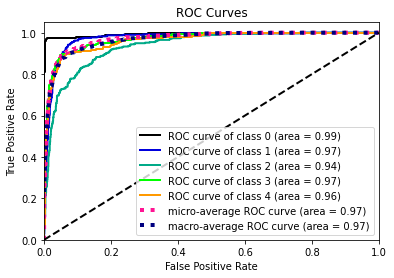}}
\caption{Receiver-operator characteristic (ROC) curves for the testing results of the best 5-class classifier.}
\label{fig}
\end{figure}

\section{Discussion}

The results show that ResNet50 and InceptionResNetV2 have relatively better performance than the other evaluated models. A commonality between these models is the implementation of skip connections. Furthermore, using the default input size for a model also yields better results.

Firstly, the implementation of skip connections may contribute to better performance as it may help the model avoid the vanishing gradient problem (VGP). During training, back-propogation may result in vanishing gradients, especially in the earlier model layers. This could impact the overall ability for the model to learn as learning could slow down in those layers, hindering the model's performance potential. Skip connections may avoid the VGP as they enable the model to flow gradients between non-consecutive layers, thus skipping over layers that may include vanished gradients. Secondly, the use of model-specific default input sizes may contribute to better performance because changing the size of the input forces the model to learn from scratch since pre-trained parameters' sizes may not match the new input layer.

\section{Conclusion and Future Work}

Promising methods for binary, 3-class, and 5-class classification of DR have been demonstrated. Additional work can be conducted, especially for the multiclass classification tasks. Overall, it was found that, given the parameters of the conducted experiments, the ResNet50/Adam combination is best for the binary task, and the InceptionResNetV2/Adam combination is best for the 3-class and 5-class tasks.

\subsection{Model Performance Improvement}

In future research, other deep learning architectures and optimization techniques could be experimented with. Furthermore, further tuning of the learning rate of the optimizer could be conducted. A hybrid approach to layer freezing could also be taken, resulting in a mix of frozen and unfrozen layers. Lastly, based on observed results, techniques explored with the multi-class task could be applied to the binary task in order to improve the results. However, high accuracy 5-class classification is the ultimate goal as it provides the most insight into the severity of the disease in an image to a medical professional, so more effort should be put towards it.

\subsection{Implementation of a Multi-Stage Classification System}

A multi-stage classification system could be developed through which a series of binary classifications in a decision tree-like manner could lead to a 5-class classification. This could be a promising approach for DR detection as binary classification has already shown extremely promising results. Sigmoid outputs of each binary model within the system would individually contribute to the final 5-prediction array, to which Argmax would be applied in order to determine the final classification by identifying the node with the highest value, which would be indicative of the image classification with the highest probability.

\subsection{Deployment of a Deep Learning-Based Medical Diagnostic Tool}

There is a strong need for a reliable diagnostic tool for the detection of diabetic retinopathy given the global prevalence of the disease. Additionally, global regions with a lack of medical professionals could significantly benefit from such a tool. After undergoing clinical trials and testing, a cost-friendly deep learning-based diagnostic tool could be deployed through a cloud-based service in order to provide global access to a reliable and accurate detection system.

\section*{Acknowledgment}

We would like to acknowledge Google Colaboratory for providing a high RAM and fast GPU environment to run deep learning experimentation in a time-efficient manner.

\end{document}